\documentclass[11pt]{article}

\usepackage[margin=1in]{geometry}
\usepackage{amsmath,amssymb,amsthm}
\usepackage{booktabs}
\usepackage{array}
\usepackage{enumitem}
\usepackage{float}
\usepackage{microtype}
\usepackage{xcolor}
\usepackage{graphicx}
\usepackage{listings}
\usepackage[most]{tcolorbox}
\definecolor{newblue}{rgb}{0.19,0.55,0.91}
\definecolor{promptbackground}{RGB}{247,247,247}
\definecolor{promptborder}{RGB}{72,72,72}
\definecolor{prompttitle}{RGB}{64,64,64}
\usepackage[colorlinks=true,citecolor=newblue,linkcolor=blue,urlcolor=blue!60!black]{hyperref}

\newcommand{\systemname}{{\mdseries\textsc{Colosseum}}}
\newcommand{\fullsystemname}{{\mdseries\textsc{Stellar Colosseum}}}

\newcolumntype{L}[1]{>{\raggedright\arraybackslash}p{#1}}
\newtcblisting{promptbox}[1]{
  enhanced jigsaw,
  breakable,
  listing only,
  colback=promptbackground,
  colframe=promptborder,
  colbacktitle=prompttitle,
  coltitle=white,
  fonttitle=\bfseries,
  title={#1},
  title after break={#1 (continued)},
  boxrule=0.7pt,
  arc=1.2mm,
  outer arc=1.2mm,
  left=1.5mm,
  right=1.5mm,
  top=1mm,
  bottom=1mm,
  before skip=8pt,
  after skip=10pt,
  listing options={
    basicstyle=\ttfamily\footnotesize,
    breaklines=true,
    breakatwhitespace=false,
    columns=fullflexible,
    keepspaces=true,
    showstringspaces=false,
    tabsize=2
  }
}

\title{\fullsystemname{}: A Many-Agent Harness for Long-Horizon Research in Mathematics and Theoretical Computer Science}
\author{%
Honghao Lin\textsuperscript{*,1}\quad
David P. Woodruff\textsuperscript{*,1,2}\\[0.6em]
Yuan Deng\textsuperscript{1}\hspace{1.5em}
Jieming Mao\textsuperscript{1}\hspace{1.5em}
Song Zuo\textsuperscript{1}\hspace{1.5em}
Vahab Mirrokni\textsuperscript{1}
}
\date{%
  \includegraphics[width=3.2cm]{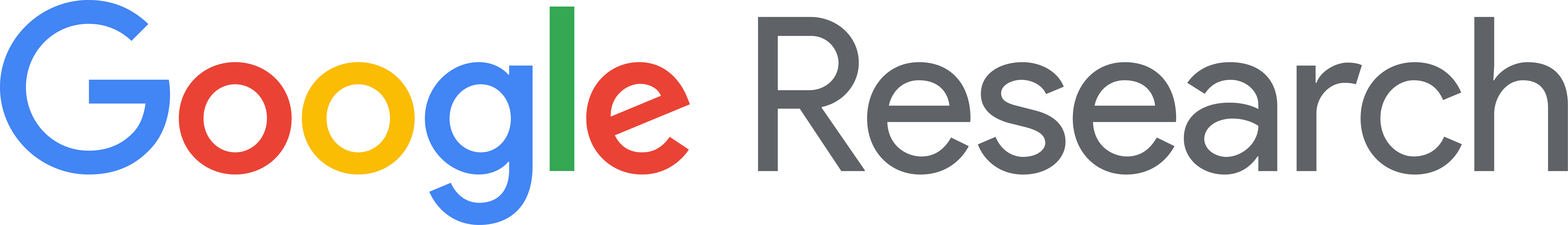}%
}

\begin{document}
\maketitle
\vspace{-1.3em} 

\begin{NoHyper}
\begingroup
\renewcommand{\thefootnote}{*}
\footnotetext[0]{Co-first authors. Email: \texttt{honghaol@google.com}, \texttt{woodruffd@google.com}.}
\endgroup
\footnotetext[1]{Google Research.}
\footnotetext[2]{Carnegie Mellon University.}
\end{NoHyper}

\begin{abstract}
Language models can produce plausible short proofs, but may still be unreliable on
long-horizon research problems, where progress depends on a sequence of uncertain
and interdependent decisions. We introduce \fullsystemname{}, a model-agnostic
harness for allocating inference across research in
mathematics and theoretical computer
science. \systemname{} explores alternative strategies before proof construction,
uses a readiness gate to decide when a route is mature enough to decompose,
represents the proof plan as interdependent section-level subproblems, and routes
verifier findings back to the affected part of the argument. Across these stages,
it generates candidates in parallel, attacks them with targeted falsification,
and combines candidates and their critiques into a single research artifact
through overlapping random-sample tree aggregation.
The \systemname{} workflow has been integrated into Google
Antigravity's Teamwork framework as the Long Proof pattern
\cite{antigravity2026teamwork}.

We demonstrate the capabilities of \systemname{} through open-ended research and
evaluations on theorem-proving and competitive programming benchmarks. Using \systemname{}
with Gemini 3.1 Pro, we obtain several new results that address open problems arising
from papers published at top venues such as FOCS and JMLR. On \textsc{TCS-Bench}
\cite{cohenaddad2026tcsbench}, a benchmark of research-level theorem-proving tasks
drawn from papers published at FOCS, STOC, and SODA, \systemname{} achieves
$71.0\%$ accuracy using Gemini 3.1 Pro and Gemini 3.7 Flash. In a separate
Codeforces evaluation using Gemini 3.1 Pro, the proof-oriented pipeline with
execution feedback solves 218 of 222 problems.
\end{abstract}

\begin{figure}[!t]
\centering
\includegraphics[width=\linewidth]{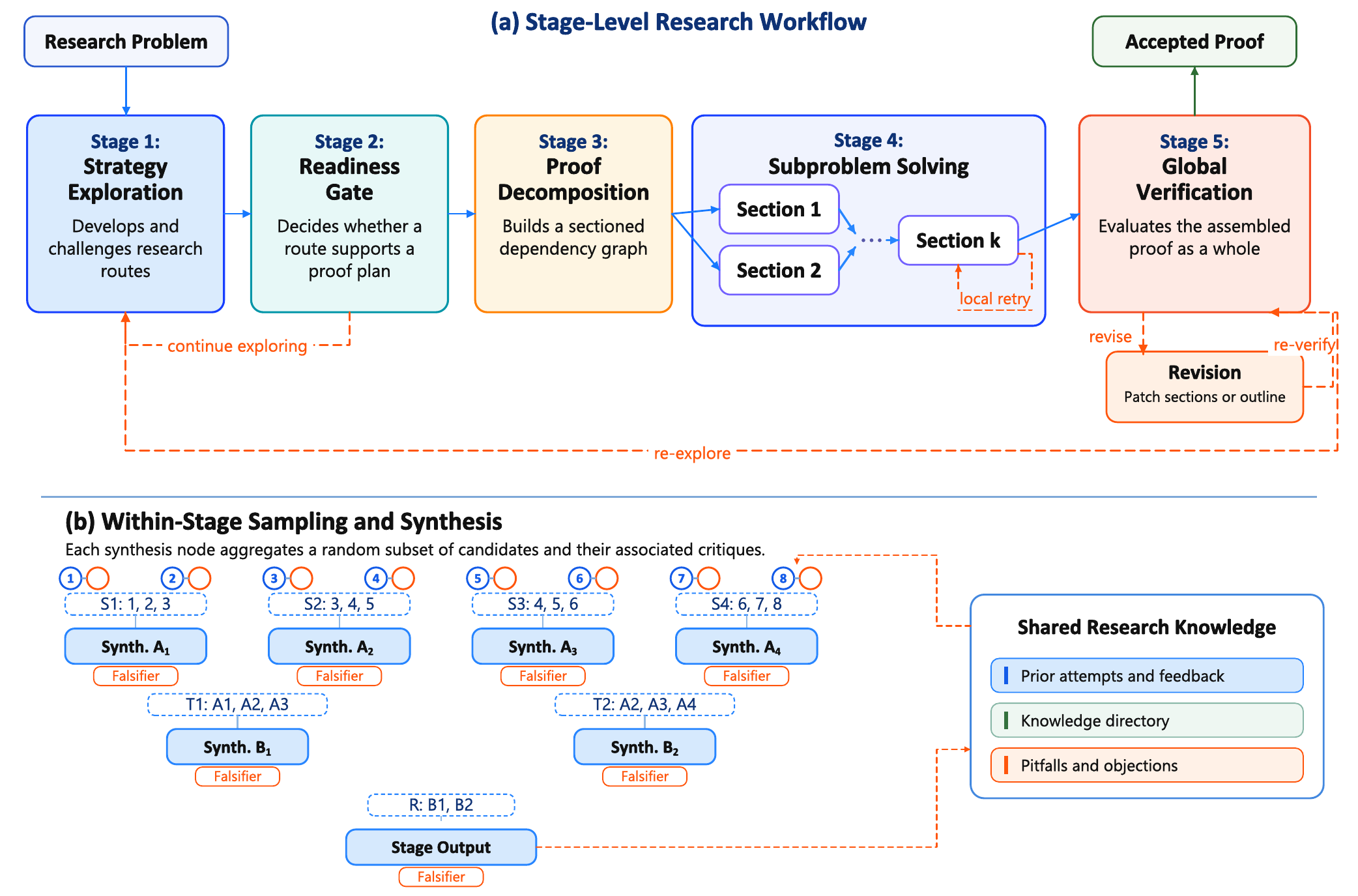}
\caption{Overview of the \systemname{} architecture. Panel~(a) shows the stage-level
workflow from strategy exploration through global verification, including local
revision and re-exploration. Panel~(b) expands inference within a stage. Synthesis
nodes aggregate random subsets of candidates and their associated critiques, and
each is paired with a falsifier.}
\label{fig:lifecycle}
\end{figure}

\section{Introduction}
\label{sec:introduction}

Language models have made substantial progress in mathematical reasoning. Systems
built around them have solved olympiad-level problems
\cite{trinh2024alphageometry,hubert2026alphaproof}. Recent studies also describe
how language models have helped obtain new results in mathematics and theoretical
computer science
\cite{feng2026aletheia,woodruff2026accelerating}. These successes motivate a
closer look at how to organize research over many rounds of exploration, proof
construction, and revision.

Work on a research problem often starts with a literature search and small
examples. Researchers may try several representations and conjectures before
finding lemmas worth proving. Even a promising approach can fail if a key lemma
turns out to be false or a later step exposes a missing assumption. As the proof
grows, definitions and assumptions must remain consistent across dependent lemmas;
changing one part may alter what later sections need to establish. A failed
approach may still yield a useful restriction, counterexample, or alternative
formulation. Progress depends on deciding what to develop, what to repair, and
when to change direction, using the evidence gathered along the way.

Inference-time scaling gives a system more opportunities to explore possible
solutions and check them. Sampling multiple reasoning paths, searching over
intermediate states, and iteratively critiquing an answer can improve reasoning
quality \cite{wang2022selfconsistency,yao2023tree,madaan2023selfrefine}. Agreement
among candidates can still hide a shared error, so a flat vote over final answers
offers limited guidance for proofs. The challenge is to use this computation to
choose between approaches, uncover specific gaps, and preserve what remains useful
after an attempt fails.

We introduce \fullsystemname{} (\systemname{} for short), a model-agnostic system
for long-horizon research in mathematics and theoretical computer science.
\systemname{} builds on and substantially extends the parallel exploration
and iterative verification architecture described by
Woodruff et al.~\cite{woodruff2026accelerating}, adding explicit control
over research stages, dependency-aware proof construction, and aggregation
that preserves critiques. The system keeps track of proposed
strategies, partial proofs, and verification findings as the work proceeds. This
shared research state allows exploration, proof construction, and revision to build
on one another.

\systemname{} organizes inference at two levels (Figure~\ref{fig:lifecycle}). At the
workflow level, it develops and challenges alternative strategies until one is concrete enough to decompose
into a sectioned plan with explicit dependencies. Independent subproblems can then
be developed in parallel, and the completed sections are assembled into a single
proof. Verification feedback guides repairs to the argument and reopens exploration
when the underlying strategy is no longer viable.

At the stage level, \systemname{} generates a population of candidate artifacts
for tasks such as strategy exploration, proof planning, and proof construction.
It subjects these candidates to targeted falsification and combines them through
overlapping random-sample tree aggregation. Critiques
remain attached to the proposals they address, so aggregation can combine useful
ideas while retaining objections and evidence of failure. The resulting artifact,
together with any unresolved objections, becomes part of the research state used
in later stages.

\paragraph{Our Contributions.}
In this paper, we make the following four contributions.
\begin{enumerate}
    \item We formulate long-horizon research as a pipeline that separates strategy
    exploration, proof decomposition, subproblem solving, and global verification.
    Decomposition produces a sectioned plan with explicit dependencies. Independent
    subproblems can then be solved in parallel; rejected proofs are revised or
    returned to exploration.

    \item We develop a stage-level adversarial inference procedure. It combines
    diverse candidate generation, targeted falsification,
    and overlapping tree-structured aggregation while retaining both synthesized
    content and evidence against it.

    \item We demonstrate \systemname{} on open-ended research problems using
    Gemini 3.1 Pro as the base model. The workflow contributed to several research
    results reported in companion papers, and we present case studies on long-form
    proof construction and independent strategy rediscovery.

    \item We evaluate \systemname{} on research-level theorem proving and
    competitive programming. On \textsc{TCS-Bench}
    \cite{cohenaddad2026tcsbench}, critique-based selection between runs using
    Gemini 3.1 Pro and Gemini 3.7 Flash reaches $71.0\%$ accuracy. In the
    competitive-programming case study, the proof-oriented workflow augmented
    with implementation and execution feedback records 218 accepted solutions
    on a 222-problem Codeforces suite.

\end{enumerate}

Notably, the \systemname{} workflow has since been integrated into Google
Antigravity's Teamwork framework as the Long Proof pattern
\cite{antigravity2026teamwork}.

\section{Related Work}
\label{sec:related}

We review work on inference-time search and critique, discovery guided by
verifiers and evaluators, and systems that support sustained research.

\subsection{Search, Critique, and Aggregation}

Sampling diverse reasoning paths and aggregating their answers
\cite{wang2022selfconsistency}, searching over intermediate states
\cite{yao2023tree}, revising against model-generated feedback
\cite{madaan2023selfrefine}, and organizing multi-agent debate
\cite{du2023debate} provide complementary ways to use additional inference.
Wu et al.~\cite{wu2025inferencescaling} compare inference strategies under fixed
compute budgets and introduce REBASE, which allocates tree expansions using
intermediate rewards.
Other work focuses on falsification. REFUTE evaluates whether language models
can construct counterexamples to incorrect competitive-programming submissions
\cite{sinha2025falsify}. Momus extracts key conjectures from stalled proof
attempts and asks fresh solver instances to try to prove each conjecture and
its negation. This context-detachment mechanism aims to escape solver--grader
``cognitive wells'' \cite{dang2026cognitivewell}.
\systemname{} uses search, critique, and aggregation throughout the research
workflow. Within each stage, proposals remain paired with their falsification reports during
aggregation. The resulting artifact, unresolved objections, and repair targets
remain available during strategy selection, proof construction, and verification.

\subsection{Discovery Guided by Verifiers and Evaluators}

Machine-checkable feedback provides a concrete test for proposals during search.
Learned proposals can guide symbolic deduction
\cite{trinh2024alphageometry}, while formal proof checking can guide theorem
proving \cite{hubert2026alphaproof}; executable objectives instead guide the
search for mathematical constructions
\cite{romeraparedes2024funsearch,novikov2025alphaevolve,
georgiev2025mathematicalexploration}. Recent systems extend this principle to
research-scale formalization through agentic Lean search
\cite{tsoukalas2026formalproofsearch} and retrieval-assisted proof development
\cite{ju2026rethlasarchon}. LeanMarathon uses an evolving blueprint and a proof
dependency graph to coordinate parallel development and local repair, with
changes checked through continuous integration \cite{zhang2026leanmarathon}.

These systems ultimately require a proof that passes formal checking against a
precise Lean statement
\cite{tsoukalas2026formalproofsearch,ju2026rethlasarchon,
zhang2026leanmarathon}. \systemname{} reviews provisional strategies,
intermediate claims, and natural-language proof drafts. Formal checks and
executable tests, when available, contribute evidence alongside model-generated
critiques. The workflow uses this combined evidence to revise its research state
and assess the assembled argument.

\subsection{Autonomous and Collective Research Workflows}

Research workflows must decide how to reuse intermediate work and when to ask
for further verification. Aletheia maintains
a generate--verify--revise loop over long natural-language solutions
\cite{feng2026aletheia}. Other systems assign these tasks to different roles:
QED separates decomposition, proof generation, and
verification \cite{an2026qed}; ProofCouncil iterates between an author and a
critic \cite{schmitt2026proofcouncil}; and RMA coordinates initializer,
proposer, and verifier agents through shared structured memory
\cite{zhao2026rma}. These systems use verification repeatedly as an argument
develops.

These systems also differ in how they store and share intermediate work. Danus admits
verifier-approved claims into a fact graph together with their proofs and
logical dependencies \cite{liu2026danus}. Other systems combine persistent
proof state, literature retrieval, computational or formal tools, and human
steering \cite{gong2026albilich,cao2026mechmath,zheng2026aicomath}. Their shared
state may be organized around verified claims, proof obligations, working files,
or an interactive workspace.

Some frameworks support ongoing research communities. Within Station,
heterogeneous agents choose their own directions and
publish an internal literature that later agent generations can extend
\cite{chung2026station}; in EinsteinArena, user-supplied agents
interact asynchronously through shared submissions, verifiers, and discussions
\cite{bianchi2026einsteinarena}.

Related work also studies collaboration and review across the scientific
research process. Gemini-assisted case studies identify decomposition and iterative
refinement as common techniques and also document models serving as adversarial
reviewers \cite{woodruff2026accelerating}. Co-Scientist applies
generate--critique--evolve loops to research hypotheses
\cite{gottweis2026coscientist}, while the AI Scientist systems extend such
orchestration through experimentation, analysis, and manuscript production
\cite{lu2024aiscientist,yamada2025aiscientistv2}. PAT applies inference-scaled
agents downstream to scientific review and verification \cite{jayaram2026pat}.

Aletheia and OpenAI have reported results on FirstProof
\cite{feng2026firstproof,openai2026firstproof}. OpenAI also reports ten advances
in mathematics and theoretical computer science, each accompanied by a Lean
certificate \cite{openai2026tenadvances}. Other reports describe Claude-assisted
results concerning the Riemann zeta function \cite{anthropic2026riemann},
cryptanalysis \cite{anthropic2026crypto}, and the Jacobian conjecture
\cite{alpoge2026jacobian,gao2026jacobian}. Differences in human involvement,
disclosure, and evaluation make it difficult to isolate the contribution of any
one workflow component to these outcomes.

\systemname{} combines these ideas to connect strategy exploration with the
development and repair of a long proof. Its readiness gate controls when a
route is decomposed into subproblems, and its aggregation procedure carries
critiques alongside the proposals they address. Shared drafts and research
knowledge keep that evidence available during both local revision and renewed
exploration.

\section{\systemname{} Overview}
\label{sec:overview}

The design of \systemname{} is motivated by four challenges in long-horizon
research:
\begin{itemize}
    \item \textbf{Strategic Uncertainty.} A precise problem statement may leave the
    route to a proof unclear. Finding useful representations, reductions, or
    intermediate targets often requires substantial exploration, and a promising
    route may remain uncertain until its central claims are tested.

    \item \textbf{Distributed Technical Difficulty.} A proof may contain several
    interdependent bottlenecks, from constructing a new object to establishing a
    delicate estimate. Resolving one difficulty can expose another, and a local
    revision may require changes to the surrounding argument.

    \item \textbf{Long Outputs and Error Accumulation.} Long proofs can exceed the
    output budget of a single response, leading to omitted details or unfinished
    arguments. Definitions and assumptions must also remain consistent across
    distant sections, since an unsupported step can be reused and propagate errors
    through the rest of the proof.

    \item \textbf{Failure and Partial Progress.} Failed attempts may still yield
    useful counterexamples, restrictions, or intermediate results. To guide later
    work, this information must be recorded as specific mathematical claims or
    objections; a generic failure judgment does not explain what remains valid or
    what should change.
\end{itemize}

We next outline the design of \systemname{} and how its main components work
together to address these challenges.

\subsection{Research Pipeline}

Figure~\ref{fig:lifecycle} summarizes the architecture built around these
considerations. Panel~(a) shows the overall research workflow, while panel~(b)
expands inference within a stage.
Exploration develops and tests candidate strategies until the readiness gate
selects a route that is concrete enough to support proof construction.
Decomposition turns that route into a sectioned proof skeleton and a dependency
graph over its subproblems. Eligible subproblems are then solved, with independent
sections processed in parallel. A section that fails its local review is retried
without restarting unaffected subproblems, and the completed sections are assembled
into a single proof.

The global verifier evaluates the assembled proof. A rejected proof either enters
revision or reopens exploration. Revision may replace one or more section bodies or
modify the proof outline before the argument is assembled and verified again.
Re-exploration is reserved for cases in which the current strategy no longer
supports a credible path to the target.
Section~\ref{sec:lifecycle-detail} develops these stages and return paths in detail.

\subsection{Inference within a Stage}

The research pipeline determines which task is currently being addressed, but a
difficult stage is not entrusted to a single model continuation. \systemname{}
generates a population of candidates, subjects them to targeted falsification, and
iteratively synthesizes random subsets of candidates and their associated critiques
through the overlapping sampling tree shown in panel~(b). The candidate type changes
with the stage---a strategy, a proof skeleton, a section body, a verifier assessment,
or a revision---but the inference pattern is reused.

The root artifact and any unresolved objections are returned to the research
pipeline. In this way, stage-level decisions determine where additional inference
is applied, while aggregation determines how a difficult decision within that
stage is resolved. Section~\ref{sec:adversarial-aggregation} describes this inner
procedure in detail.

\subsection{Context across Stages}

Information produced earlier in the pipeline remains available to later stages.
Within a round, this includes the current strategy, sectioned proof plan, completed
section bodies, and verifier findings. Across rounds, a rejected proof draft and
its verifier feedback are passed directly into the next attempt. A shared knowledge
directory further retains reusable results from the wider search.
Section~\ref{sec:context} describes these two forms of cross-round memory.

\section{Architecture and Workflow}
\label{sec:architecture-workflow}

\subsection{Adversarial Generation and Tree-Structured Aggregation}
\label{sec:adversarial-aggregation}

At a difficult stage, \systemname{} generates a population of candidates, subjects
each candidate to targeted falsification, and reduces the resulting
candidate--critique bundles through a tree. Reviewer roles and output schemas vary
by stage; the three operations stay the same.

\subsubsection{Parallel Candidate Generation}

Parallel generation seeks differences that can change the mathematical outcome:
representation, principal lemma, proof technique, case split, interpretation of
evidence, or location of the main bottleneck. Seeds, temperatures, prompt
perspectives, tool access, and assumed lines of attack provide additional sources
of variation.

Each candidate follows a typed schema appropriate to its stage. A strategy
proposal, for example, states its mechanism, required lemmas, expected bottleneck,
and a falsifiable test. A proof proposal states its assumptions and marks any gaps.
The schemas make candidates comparable and give reviewers specific claims to
attack.

Selected prompt templates and their structured output interfaces are provided
in Appendix~\ref{app:prompts}.

\subsubsection{Targeted Falsification}

One or more adversarial reviewers examine each candidate. They concentrate on
finding defects and propose an alternative solution only when it helps establish
one. The review covers:
\begin{itemize}
    \item counterexamples and boundary cases;
    \item invalid implications or silently strengthened hypotheses;
    \item circularity and undeclared dependencies;
    \item misuse of a theorem, computation, or external reference;
    \item a mismatch between the proved statement and the target claim; and
    \item missing assumptions or results needed by later sections.
\end{itemize}

Falsification records remain attached to the candidate. A clean record may reflect
weak tests, whereas a precise objection can make another candidate easier to
repair. Here \emph{falsification} refers to the search objective; failure to find a
defect does not establish correctness.

\subsubsection{Tree-Structured Aggregation}

Comparing every candidate and critique in one prompt becomes unwieldy at large
sample counts. \systemname{} uses a reduction tree instead. Let \(z_i\) denote a
candidate bundled with all of its falsification records, and let
\[
  \mathcal{C}^{(0)}=\{z_1,\ldots,z_n\}.
\]
The tree shape is specified by the population width at every level,
\[
  (m_0,m_1,\ldots,m_L),\qquad m_0=n,\quad m_L=1.
\]
For the transition from level \(\ell\) to level \(\ell+1\), each of the
\(m_{\ell+1}\) aggregation nodes independently draws \(k_\ell\) distinct inputs
uniformly from the current population, where \(k_\ell\) is the configured sample
size capped by \(m_\ell\). Thus
\[
  G_j^{(\ell)} \sim
  \operatorname{Unif}\!\left\{G\subseteq\mathcal{C}^{(\ell)}:
  |G|=k_\ell\right\},
  \qquad j=1,\ldots,m_{\ell+1}.
\]
Sampling is without replacement within one group, but groups for different
aggregation nodes are drawn independently and may overlap. They do not form a
partition. Given the stage-specific aggregator \(A\), the current input artifact
\(x\), and contextual evidence \(K\), each node computes
\[
  s_j^{(\ell+1)}
  =
  A\!\left(x,K,G_j^{(\ell)}\right),
  \qquad
  \mathcal{C}^{(\ell+1)}
  =
  \{s_j^{(\ell+1)}\}_{j=1}^{m_{\ell+1}}.
\]
If \(R_i^{(\ell)}\) is the number of next-level groups containing a fixed node from
level \(\ell\), then
\[
  \mathbb{E}\!\left[R_i^{(\ell)}\right]
  = \frac{m_{\ell+1}k_\ell}{m_\ell}.
\]
Ordinary contraction layers choose widths and sample sizes that give an expected
reuse of roughly two to three. A transition from 128 nodes to 64 with sample size
five, for example, gives expected reuse 2.5. Mixing layers and the final root may
use a different rate. Overlap gives a candidate several chances to contribute while
keeping each aggregation context small.

Intermediate aggregation is constructive rather than a vote or ranking: it may
merge compatible components, retain competing branches, repair a localized flaw,
or declare an unresolved conflict. Substantive disagreements and falsification
evidence are carried forward rather than averaged away. The root returns a
synthesized candidate together with any unresolved objections.

\subsection{The Research Pipeline}
\label{sec:lifecycle-detail}

\subsubsection{Strategy Exploration and Readiness}

\systemname{} begins by exploring proof strategies rather than drafting a proof
around the first plausible idea. Parallel attempts develop routes based on
different reformulations, intermediate claims, and connections to known results.
The purpose of this stage is to expose what each route would require, which parts
remain conjectural, and where the main technical difficulties lie.

The readiness gate asks whether one of these routes is concrete enough to support a
proof plan. It tests whether the remaining uncertainty can be localized within a
stable proof architecture, not whether the proof has already been completed. A
route passes when its central reduction or mechanism is stable, its unresolved
claims are precise enough to assign to proof sections, and no unresolved bridge is
likely to change the target or the architecture of the argument. A central lemma
may remain difficult; what matters is that its statement, role, and expected path
to proof or verification are explicit. If these conditions are not met,
exploration continues; otherwise, the selected route is passed to decomposition.

\subsubsection{Decomposition and Parallel Proof Construction}

The decomposer turns the selected route into a numbered, sectioned proof skeleton.
Each section is associated with a subproblem specifying the mathematical content
that must be established there. Dependency edges record which completed sections a
subproblem may use, producing a directed acyclic graph. The document order controls
the intended exposition, while the dependency graph controls the order in which
the mathematical work can proceed.

A subproblem becomes eligible once its dependencies have been completed, so
independent sections can be solved in parallel. Each solver receives the assigned
task together with the relevant completed sections. Difficult subproblems can use
the adversarial inference procedure of Section~\ref{sec:adversarial-aggregation} to
develop and test alternative solutions.

Before a section is committed to the skeleton, a local reviewer checks it against
the assigned subproblem and the completed material on which it depends. If the
section remains incomplete or the review identifies an unresolved defect, the same
subproblem is run again with the failed section and the review as additional input.
The retry is local to the affected subproblem: completed work elsewhere in the
dependency graph is preserved. Once a section is accepted locally, its body
replaces the corresponding placeholder and becomes available to downstream
subproblems. Filling the skeleton in this way produces the candidate proof
submitted to global verification.

\subsubsection{Global Verification and Feedback}

The global verifier reads the original problem and the completed document as a
single argument, with the local subproblem reviews available as audit context. A
collection of individually plausible sections may still fail as a proof because a
dependency is used with the wrong assumptions, notation or definitions drift
across sections, a case is omitted, or the final conclusion does not match the
original target. The verifier also checks that a conditional or attacked claim from
a local review has not been silently inherited by the assembled proof.

Global verification is itself carried out by multiple independent reviews and
tree-structured aggregation. The aggregation merges genuinely duplicate
criticisms while retaining distinct substantive objections. It is not a majority
vote: a concrete fatal defect is sufficient to reject the proof, and generic
acceptance judgments do not resolve it.

The final review gives an overall verdict together with the material defects found
in the argument. Each defect is tied to the section or claim where it arises; for a
cross-section dependency error, the review identifies both the supporting section
and the section that uses it. This localization makes the global review actionable
without reducing it to a collection of independent section checks.

When the verifier rejects a proof, the pipeline proceeds by revision or
re-exploration. Revision is used when the current strategy remains viable; it may
replace section bodies or modify the proof outline before the entire argument is
verified again. If the verifier's findings undermine the central strategy, the
system returns to exploration instead.

\subsection{Shared Research Knowledge}
\label{sec:context}

\systemname{} carries information across research rounds in two complementary
forms: the latest proof attempt is retained in full, while a shared knowledge
directory collects reusable findings from the wider search. The former grounds the
next round in the argument that was actually attempted; the latter preserves useful
results that may otherwise be lost as candidates are aggregated or discarded.

\subsubsection{Retaining Prior Attempts}

When a round does not produce an accepted proof, \systemname{} passes the resulting
draft and the verifier feedback directly into the next round. Revision can address
the concrete defects identified in the draft, while re-exploration can reconsider
the strategy in light of the argument that failed. The next attempt therefore does
not begin from the original problem alone.

Retaining a draft does not endorse its claims. The verifier feedback remains
attached to it, making clear which steps or strategic assumptions require further
work.

\subsubsection{Knowledge Directory}

The latest draft is only one product of a much wider search. A knowledge curator
reads strategy proposals and falsification reports and records four kinds of reusable
knowledge:
\begin{enumerate}
    \item \textbf{theorems and lemmas}: mathematical results developed during the
    search, together with their hypotheses, supporting arguments, and possible
    applications;
    \item \textbf{failed approaches}: attempted routes, their precise failure
    points, and conditions under which a variant might still work;
    \item \textbf{references}: relevant literature, including the statements and
    hypotheses needed for the current problem; and
    \item \textbf{observations}: structural properties or computational findings,
    together with their evidence and implications for subsequent work.
\end{enumerate}
Entries retain their source and relevant caveats. The directory is updated across
research rounds and made available to subsequent agents, allowing them to reuse
earlier results and avoid repeating approaches whose failure has already been
identified.

\subsection{Inference Configurations}
\label{sec:inference-configurations}

In this section, we give the inference configurations used in our research
campaigns and evaluations. For open-ended research problems
(Section~\ref{sec:results}), we use a range of tree configurations during strategy
exploration, some with slightly over 100 leaf nodes. These larger trees are used only
for exploration; all remaining stages share a fixed configuration with 16 leaf
nodes.

Table~\ref{tab:inference-configurations} summarizes the strategy-exploration
configurations for open-problem research, \textsc{TCS-Bench}
(Section~\ref{sec:tcsbench}), and Codeforces (Section~\ref{sec:codeforces}), together
with the shared configuration used by all remaining stages. Tree widths and
per-node sample sizes follow the notation in
Section~\ref{sec:adversarial-aggregation}.

\begin{table}[H]
\centering
\small
\begin{tabular}{llcc}
\toprule
Setting & Stage & Tree widths \(\mathbf{m}\) & Sample size \(k\) \\
\midrule
Open-problem research
    & Strategy exploration & Varies & Varies \\
\textsc{TCS-Bench}
    & Strategy exploration & \((32,16,8,5,1)\) & \(5\) \\
Codeforces
    & Strategy exploration & \((32,16,8,5,1)\) & \(5\) \\
All three settings
    & All remaining stages & \((16,8,5,1)\) & \(5\) \\
\bottomrule
\end{tabular}
\caption{Stage-level aggregation configurations for open-problem research,
\textsc{TCS-Bench}, and Codeforces. At level \(\ell\), the effective sample size
is \(\min\{k,m_\ell\}\).}
\label{tab:inference-configurations}
\end{table}

These configurations specify population widths and aggregation fan-in rather than
the exact total number of model calls, which also depends on the number of proof
sections, local retries, and global revision rounds.

\section{Selected Research Results}
\label{sec:results}

We summarize five research results to which runs of the workflow contributed.
Each subsection states the motivating question and principal advance; the cited
papers provide complete definitions, attribution, and proofs.

\subsection{Strong Coresets for \texorpdfstring{$\ell_p$}{lp} Subspace Approximation When \texorpdfstring{$p>2$}{p greater than 2}}

Given a matrix $A\in\mathbb{R}^{n\times d}$, the $\ell_p$ subspace approximation
problem asks for a low-dimensional subspace that minimizes the aggregate distance
of the rows of $A$ to that subspace. A strong coreset samples and rescales a small
number of rows to form $SA$ while preserving, simultaneously for every subspace
$F$ of dimension at most $k$, the objective
\[
  \lVert SA(I-P_F)\rVert_{p,2}^p
  = (1\pm\varepsilon)\lVert A(I-P_F)\rVert_{p,2}^p.
\]
For $p>2$, Woodruff and Yasuda obtained coreset size
$\widetilde{O}_p(k^{p/2}\varepsilon^{-p})$ using a recursive ridge-leverage-score
sampling framework \cite{woodruff2024ridge}. It was unclear whether the
$\varepsilon^{-p}$ dependence was intrinsic to the sampling rule or arose from
the analysis of the recursive row-count recurrence.

The proof initially obtained by \systemname{} shows that the same sampling rule supports
a coreset of size $\widetilde{O}_p(k^{p/2}\varepsilon^{-2})$, while retaining
$\widetilde{O}_p(\operatorname{nnz}(A)+d^\omega)$ running time. The key step keeps
the truncation in the sampling probabilities when bounding the surviving rows,
which changes the fixed point of the recurrence from an $\varepsilon^{-p}$ to an
$\varepsilon^{-2}$ dependence. The complete construction and analysis are given
in \cite{lin2026coresets}.

\subsection{The Condition-Number Barrier in Sparse Least Squares}

Sparse convex optimization seeks a vector with few nonzero coordinates whose
objective value is close to that of the best $k$-sparse solution. For objectives
with restricted condition number $\kappa$, known algorithms require output
sparsity with essentially linear dependence on $\kappa$. Axiotis and Sviridenko
gave an algorithm achieving this dependence and conjectured that it could not be
improved by a polynomial-time algorithm \cite{axiotis2021sparse}.

For least-squares objectives, the new result establishes this barrier conditional
on the randomized exact-volume Small-Set Expansion Hypothesis. Under this
hypothesis, for every fixed $\gamma\in(0,1]$, no randomized polynomial-time
algorithm can, with probability
at least $2/3$, return a vector $x$ such that, writing $s=\lVert x\rVert_0$,
\[
  \lVert Ax-b\rVert_2^2
  \leq \min_{\lVert z\rVert_0\leq k}\lVert Az-b\rVert_2^2+\varepsilon
  \quad\text{and}\quad
  s=O\!\left(k\kappa_{s+k}^{1-\gamma}\right),
\]
where $\kappa_r$ is the restricted condition number at sparsity level $r$.
Thus, no fixed sublinear power of the condition number can replace the linear
dependence. The complete reduction and parameter
regime are given in \cite{lin2026condition}.

\subsection{Dimension Lower Bounds for Maximum Inner Product Embeddings}

Multi-vector embeddings represent an item by a point cloud and compare point
clouds using Chamfer similarity, whereas single-vector embeddings compare one
vector per item by an inner product. For singleton queries, Chamfer similarity
reduces to maximum inner product similarity. An upper bound of
$m^{O(1/\varepsilon^2)}$ was known for representing point clouds of size at most
$m$ by single vectors, while the earlier lower bound
$(\varepsilon^2m)^{\Omega(1/\varepsilon)}$ left a gap between
$1/\varepsilon$ and $1/\varepsilon^2$ in the exponent of $m$
\cite{jayaram2026multivector}.

The new lower bound nearly closes this gap. For every fixed
$\delta\in(0,1)$ and sufficiently large $m$, there are unit query vectors and
document point clouds for which any single-vector representation approximating
all maximum inner products to additive error $\varepsilon$ must have dimension
\[
  D \geq m^{c_\delta/\varepsilon^{2-2\delta}}
\]
for a constant $c_\delta>0$. The lower bound holds even for fully data-dependent
representations and extends to Chamfer similarity. The construction and
approximate-rank argument are given in
\cite{jayaram2026maxip}.

\subsection{Single-Stage Hadamard Quantization}

Randomized Hadamard transforms provide fast preprocessing for quantizing
high-dimensional vectors in similarity search, distributed learning, and model
compression. Earlier work gave an unbiased dithered quantizer with sharp
mean-squared error guarantees \cite{feng2026quantization}. Its finer
$1/d$-scale inner-product estimator, however, used a second randomized transform
and a residual quantization stage, adding both communication and a larger leading
constant.

The new estimator shows that the second stage is unnecessary for attaining the
same $1/(d4^b)$ mean-squared-error scaling. Pairwise-independent dithers across
Hadamard coordinates yield an unbiased, single-stage estimator using $b$ bits per
coordinate and satisfying
\[
  \mathbb{E}\!\left[
    \left|\left\langle y,\widehat{x}-x\right\rangle\right|^2
  \right]
  \leq
  \left(\frac{3\pi\sqrt{3}}{2}+o(1)\right)
  \frac{\lVert y\rVert_2^2}{d\,4^b}.
\]
It removes the residual-stage $O(d)$-bit payload and reduces the leading constant
in the proved upper bound by a factor of approximately $5.93$. The estimator and
proof are given in \cite{lin2026dithering}.

\subsection{Lower Bounds for Prefix-Matrix Factorizations}

Let $Q$ be the $n\times n$ lower-triangular all-ones matrix, which maps a vector to
its sequence of prefix sums. For a factorization $Q=AB$, the quantity
\[
  \gamma_{2,1}(Q)
  = \inf_{Q=AB}
    \lVert A\rVert_{2\to\infty}\lVert B\rVert_{1\to1}
\]
governs space bounds for factorization-based rank and quantile estimation in
turnstile streams, as well as error bounds for matrix mechanisms in continual
counting. The matrix-factorization framework in \cite{bulanek2026matrix} made a
sharp characterization of this cost a central question.

The new result proves the near-optimal lower bound
\[
  \gamma_{2,1}(Q)
  = \Omega\!\left(
      \frac{\log^{3/2}n}{(\log\log n)^{3/2}}
    \right)
\]
over real factorizations of arbitrary finite inner dimension. A dyadic
factorization gives the upper bound $O(\log^{3/2}n)$, so the two bounds differ by
only a factor of $(\log\log n)^{3/2}$. The proof combines right-sided Haar
projections with a scale-dependent numerical-sparsity
decomposition, then aggregates the resulting estimates across dyadic scales. The
full lower bound is given in \cite{lin2026prefix}.

The following two case studies complement the results above by illustrating the
workflow on unusually long proof artifacts and under information isolation.

\subsection{Case Study: Long-Form Proof Construction for Knuth's Cycles}

For an integer $m>2$, Knuth's cycles problem asks whether the directed edges of
the Cayley graph
\[
  \Gamma_m=\operatorname{Cay}\!\left(\mathbb{Z}_m^3,
  \{e_1,e_2,e_3\}\right)
\]
can be partitioned into three directed Hamiltonian cycles. The odd case admits a
simple construction with a complete proof, while the even case remained more
difficult \cite{knuth2026claudecycles}. The updated notes also record an earlier,
more intricate even-case construction generated by GPT-5.3-Codex, for which a
complete proof was subsequently obtained with GPT-5.4 Pro
\cite{gpt54pro2026evenproof}. A later
multi-agent search produced a much simpler construction for even $m$ and verified
it computationally through $m\leq 2000$
\cite{aquinomichaels2026completing}, but a
rigorous symbolic proof for arbitrary even $m$ remained open. A subsequent note
introduced a second simple construction and reports full-length AI-generated proof
drafts for both \cite{brenner2026knuthcycles}.

Although each construction is specified by a compact set of local routing rules,
proving that it yields three Hamiltonian cycles requires a substantially more
involved global argument. The rules depend on parity and include several
exceptional boundary cases, whose interactions must be controlled uniformly for
arbitrary even $m$. In particular, the proof must show that every directed edge is
assigned exactly once and that each color class traverses all $m^3$ vertices in a
single cycle rather than decomposing into shorter cycles. The workflow then
produced a 46-page proof draft for the earlier construction and a 75-page proof
draft for the new one \cite{brenner2026knuthcycles}. This case study shows how a
proof far beyond the length of a typical single model response can instead be
developed as a persistent, revisable document organized around an explicit
dependency structure.

\subsection{Case Study: Independent Rediscovery of the Erdős Unit-Distance
Breakthrough}

Let $u(n)$ be the maximum number of unit-distance pairs determined by $n$ points
in the plane. Erdős's classical grid construction gives
$n^{1+\Omega(1/\log\log n)}$ unit-distance pairs, and he conjectured that
$u(n)=n^{1+o(1)}$. OpenAI reports that an internal model generated a
counterexample to this long-standing conjecture in 2026. A group of
mathematicians then distilled, simplified, and human-verified the AI-generated
argument, establishing that infinitely many point sets determine at least
$n^{1+\delta}$ unit-distance pairs for some fixed $\delta>0$
\cite{alon2026unitdistance}.

We then ran \systemname{} on the same problem using Gemini 3.1 Pro as the base
model, with internet access disabled. The resulting 22-page draft
\cite{colosseum2026erdos}, available on
\href{https://github.com/dpwoodru/erdos}{GitHub}, developed a number-theoretic
approach based on unramified towers and relative unit groups, independently
arriving at the central architecture of the OpenAI solution
\cite{alon2026unitdistance} and pursuing a bound of the form
$u(n)\geq n^{1+\varepsilon_0}$ for infinitely many $n$, with
$\varepsilon_0>0$ given explicitly. Notably, the run proceeded through 15
exploration rounds. Across these rounds, the accumulated-knowledge layer carried
partial results, objections, and failed attempts forward across successive rounds,
allowing the strategy-selection loop to synthesize the accumulated evidence into
subsequent research directions. This provides a concrete example of the
architecture sustaining a coherent long-horizon research process.

\section{Research-Level Evaluation on TCS-Bench}
\label{sec:tcsbench}

The preceding results and case studies examine individual problems in depth.
To evaluate the breadth of the system on a common set of research-level problems,
we also test \systemname{} on \textsc{TCS-Bench} \cite{cohenaddad2026tcsbench}.
The benchmark contains 300 theorem-proving tasks derived from papers published at
FOCS, STOC, and SODA between 2020 and 2026. Each task supplies the mathematical
context needed to state the problem and asks the model to produce a self-contained
proof of a target theorem. Candidate proofs are scored by a reference-assisted
automated grader that also receives the benchmark's ground-truth proof. Its prompt
was optimized on a separate set of 100 expert-labeled proofs, on which it reported
more than $90\%$ accuracy. All benchmark accuracies below are measured by this
grader.

We run \systemname{} separately with Gemini 3.1 Pro and Gemini 3.7 Flash, producing
one candidate proof from each run for every problem. To select between the two
candidates, Gemini 3.7 Flash produces eight independently sampled critiques of the
Gemini 3.1 Pro proof. If at least five critiques judge that proof correct, it is
submitted; otherwise, the Gemini 3.7 Flash proof is submitted. The benchmark grader
is used only to score the selected proof and plays no role in the selection rule.

\begin{table}[H]
\centering
\begin{tabular}{lc}
\toprule
Method & Accuracy \\
\midrule
\multicolumn{2}{l}{Direct model evaluation} \\
Gemini 3.1 Pro            & $30.3\%$ \\
Gemini 3.1 DeepThink      & $52.0\%$ \\
GPT-5.6 Pro (max)         & $68.0\%$ \\
\midrule
\multicolumn{2}{l}{\systemname{} evaluation} \\
\systemname{} with Gemini 3.1 Pro   & $54.0\%$ \\
\systemname{} with Gemini 3.7 Flash & $55.0\%$ \\
\midrule
Cross-model selection & $71.0\%$ \\
Oracle best-of-two     & $77.3\%$ \\
\bottomrule
\end{tabular}
\caption{Direct-model baselines, two \systemname{} runs, and cross-model
selection on \textsc{TCS-Bench}. The oracle reports the fraction of problems
solved by at least one run and is an upper bound rather than an achievable
selection rule.}
\label{tab:tcsbench}
\end{table}

The two individual runs have nearly identical overall accuracy, but their errors
are sufficiently complementary for cross-model selection to solve 213 problems,
an improvement of 48 problems over the stronger individual run. The critique
signal distinguishes proofs labeled correct and incorrect by the benchmark grader
with an AUC of $0.896$; routing with Gemini 3.1 Pro's internal verifier alone yields
$64.7\%$ accuracy. Among the evaluated non-oracle methods on this dataset,
cross-model selection gives the highest accuracy.

\section{Case Study: Competitive Programming}
\label{sec:codeforces}

To test whether the same architecture transfers from theorem proving to executable
algorithmic tasks, we follow the Codeforces evaluation described for Gemini 3 Deep
Think: all 222 problems with \href{https://clist.by}{clist.by} difficulty estimates
above 1500 from contests held between April and October 2025
\cite{googledeepmind2026deepthink}. In our evaluation snapshot, these problems come
from 52 \href{https://codeforces.com}{Codeforces} contests numbered 2084--2162.
For the rating calculation, we use the numerical difficulty estimates stored in
the evaluation corpus. These values are on a Codeforces-like scale and are
distinct from the problem ratings displayed by Codeforces. They have a median of
2381 and range from 1530 to 4599; Table~\ref{tab:codeforces-distribution} gives the
full distribution.
Each submitted C++ solution is compiled and run against the complete hidden test
set using the problem's original checker, and is accepted only if every test
passes. Strict as-submitted grading yields the same score, so no accepted solution
depends on an output repair.

\begin{table}[H]
\centering
\small
\begin{tabular}{lcccccc}
\toprule
Estimated rating band
  & $<2000$ & $[2000,2400)$ & $[2400,2800)$
  & $[2800,3200)$ & $[3200,3400)$ & $\geq 3400$ \\
\midrule
Problems & 59 & 53 & 44 & 32 & 8 & 26 \\
\bottomrule
\end{tabular}
\caption{Distribution of the evaluation-corpus difficulty estimates for the 222
Codeforces problems.}
\label{tab:codeforces-distribution}
\end{table}

Previous evaluations map model performance onto the Codeforces scale using either
contest-calibrated rating systems or probabilistic models over rated problem sets
\cite{quan2025codeelo,zheng2025livecodebenchpro,zhou2026opendeepthink}. Because
the resulting number depends on both the corpus and the calibration rule, we define
ours explicitly and do not treat it as an official contestant rating. A problem
with estimated difficulty $r$ is treated as an opponent that a contestant of
strength $x$ solves with probability
\[
  P(\text{solve}\mid r,x)=\frac{1}{1+10^{(r-x)/400}}.
\]
For difficulty estimates $r_1,\ldots,r_{222}$, the corpus-level performance
rating $\hat{x}$ is the unique solution to
\[
  \sum_{i=1}^{222}\frac{1}{1+10^{(r_i-\hat{x})/400}}
  =n_{\mathrm{solved}}.
\]

Competitive-programming systems illustrate two ways of scaling inference.
AlphaCode and AlphaCode 2 generate large candidate populations and reduce them
through compilation, sample filtering, and behavior-based selection
\cite{li2022alphacode,alphacodeteam2023alphacode2}. Later systems make execution
iterative: tool-assisted reasoning, differential or generated tests, and repair
turn runtime evidence into feedback for the next attempt
\cite{elkishky2025competitive,li2026grandcode,
wang2026cpagent,li2026solvita}. The case study below asks a different systems
question: whether a proof-oriented exploration--decomposition workflow transfers
to executable programs without being replaced by a code-specific controller.

\paragraph{Architecture Transfer.}
When applied to competitive programming, \systemname{} retains the proof-oriented
architecture used for mathematical research. Exploration searches over solution
strategies and passes the selected route to the decomposer. Rather than
partitioning a program into software components,
the decomposer represents that route as a directed acyclic graph of mathematical
and algorithmic subproblems. Each subproblem is solved and aggregated as described
in Section~\ref{sec:architecture-workflow}. The decomposition differs only at its
endpoint: a terminal implementation subproblem realizes the algorithm established
by the preceding nodes as a single C++ program.
Table~\ref{tab:codeforces-2084f-decomposition} gives an example decomposition for
Codeforces 2084F.

\begin{table}[H]
\centering
\small
\begin{tabular}{cL{0.68\linewidth}c}
\toprule
Node & Subproblem & Depends on \\
\midrule
1 & Equivalence of Reachability to Inversion Subset & --- \\
2 & DAG Bounds Propagation using Fenwick Trees & 1 \\
3 & EDF Scheduling of Missing Elements & 2 \\
4 & Global Verification via Fenwick Trees & 3 \\
5 & \textbf{C++ Implementation} & 1, 2, 3, 4 \\
\bottomrule
\end{tabular}
\caption{Dependency graph produced for Codeforces 2084F.}
\label{tab:codeforces-2084f-decomposition}
\end{table}

\paragraph{Execution Feedback.}
The one addition to this reasoning pipeline is a C++ execution probe. Once a
candidate implementation is available, the probe compiles and runs it on public
samples and model-generated stress inputs, returning checker outcomes together
with time and memory measurements to the existing verification--revision loop.
Hidden tests remain inaccessible to the workflow and are used only for final
grading. The probe makes implementation-stage failures available to the existing
verification--revision loop.

\paragraph{Results.}
With Gemini 3.1 Pro as the base model, the execution-enabled configuration
records 218 accepted solutions on the 222-problem suite and obtains a corpus-level
performance rating of 4263. The comparison configuration without the execution
probe records 213 accepted solutions and obtains a rating of 3918.
These results show that the same proof-oriented architecture used for mathematical
research can also achieve a high acceptance rate on competitive-programming
problems.

\begin{table}[H]
\centering
\begin{tabular}{lcc}
\toprule
Configuration & Accepted & Performance rating \\
\midrule
Without execution probe & 213 & 3918 \\
With execution probe    & 218 & 4263 \\
\bottomrule
\end{tabular}
\caption{Comparison of the baseline and execution-enabled configurations on the
222-problem Codeforces evaluation.}
\label{tab:codeforces-ablation}
\end{table}

\section{Conclusion and Future Directions}
\label{sec:future-directions}

We presented \systemname{}, a model-agnostic many-agent harness for long-horizon
research in mathematics and theoretical computer science. It connects strategy
exploration, proof construction, and revision through a shared research state,
while tree aggregation combines candidates and their critiques within each stage.
The harness contributed to several new research results and achieved strong
performance on \textsc{TCS-Bench} and in competitive programming.

We describe two classes of possible extensions to \systemname{} that operate at
different timescales. Within a run, the workflow could adapt the organization of
exploration, the local dependency structure, and the allocation of inference
compute to the evolving state of the investigation. Across runs, validated
research trajectories could be used as post-training data to improve the
underlying model.

\subsection{Adapting the Inference-Time Workflow}

\paragraph{Clustered Exploration of Distinct Research Directions.}
Exploration currently aggregates a mixed population of candidate strategies. When
many candidates develop variants of the same idea, a less common but genuinely
different direction may disappear before it has been explored in sufficient depth.
An extension is to cluster strategies by their central mechanism, reduction, or
representation, and to run a separate exploration--falsification--aggregation
process within each cluster. The resulting cluster-level strategies can then be
compared and integrated at a later stage.

Preliminary, non-systematic runs suggest that this organization can preserve
productive minority directions and produce viable inputs to downstream proof
construction. More systematic work is needed to determine how clusters should be
formed and updated, how inference should be allocated among them, and when
distinct directions should be merged.

\paragraph{Local Restructuring after Subproblem Failure.}
The current local retry keeps the assigned subproblem and its dependencies fixed
while regenerating the corresponding section. Repeated failure may instead indicate
that the subproblem boundary is poorly chosen: the task may combine several
distinct claims, depend on an intermediate result that was not isolated, or require
a different interface with nearby sections.

A local restructuring step could revise a small neighborhood of the dependency
graph before retrying. It might split the failed subproblem, change the tasks of
adjacent sections, introduce an intermediate section, or redirect local dependency
edges. The revised neighborhood would retain its interface with the rest of the
proof and remain acyclic, so completed sections outside that neighborhood would not
need to be regenerated. This would provide an intermediate response between retrying
one fixed task and revising the proof outline as a whole.

\paragraph{Adaptive Inference Allocation.}
The current configurations fix population widths, aggregation fan-in, and sample
counts before a run begins. The value of an additional sample or round, however,
can vary across stages and subproblems. A future controller could use signals
already produced by the workflow, such as strategy diversity, unresolved
objections, repeated local failures, and agreement among independently generated
candidates, to expand uncertain branches, grant additional retries to unstable
sections, and stop stages whose outputs have stabilized. Compute-matched
evaluation would be needed to distinguish improved allocation from simply using
more inference.

\subsection{Learning from Research Trajectories}

The harness records structured research trajectories rather than only final
answers: candidate strategies, falsifier critiques, aggregation decisions,
dependency graphs, intermediate drafts, and revision histories. A natural
direction is to use trajectories from runs with externally validated outcomes as
post-training data for the base model. Intermediate states could provide
supervision for strategy selection, decomposition, objection handling, and
revision, while rejected routes and verifier feedback could supply negative and
corrective signals. This could distill some of the benefits of inference-time
orchestration into the underlying model and improve the starting point for later
runs. The main challenge is credit assignment: a successful final result does
not by itself reveal which intermediate strategies, critiques, or revisions were
responsible for progress. The branching structure of the harness may help
identify useful training signals by comparing candidates that share the same
context but lead to different downstream outcomes.

\bibliographystyle{plain}
\bibliography{references}

\clearpage
\appendix
\section{Selected Prompt Templates and Information Flow}
\label{app:prompts}

The templates below are shortened versions of the prompts used by the harness.
They show what each stage does, what information it receives, and what it passes
to later stages. Repeated instructions and implementation details are omitted.
Items in braces are filled in by the harness at runtime.

\subsection{Strategy Exploration}
\label{app:prompt-exploration}

Strategy exploration combines an explorer, a paired falser, an exploration
aggregator, and a readiness gate. The gate approves a strategy for decomposition,
allows a stable route to proceed with explicit proof obligations, or returns it
for further exploration.

\begin{promptbox}{Explorer}
Input:
  {problem}
  {knowledge directory, when available}
  {exploration round}
  {previous strategy, falser report, and readiness-gate report, when
   available}
  {previous proof and verifier feedback, when available}

Develop one promising high-level solution strategy before decomposition.
Return one strategy card, not a proof, section plan, or LaTeX document.
Normalize the target, state a candidate reduction and core mechanism, and
identify the first hard obstruction.

Maintain exactly one concrete primary route and at most two genuinely
distinct backup routes. Preserve an affirmative route unless a reproducible
counterexample or symbolic contradiction already refutes the target. For any
gateway test, state both the proof route if it succeeds and the
counterexample route if it fails.

Distinguish proved facts from partial, heuristic, or unverified claims.
Record risky lemmas, adversarial notes, sanity checks, and every blocking
doubt as an active obligation. Treat the strategy as ready for
decomposition only when no unresolved fatal obligation remains.

Return a strategy card that records the normalized problem, the primary
and backup routes, their supporting evidence and risky lemmas, the first
hard obstruction, active obligations, a concrete next task, and a
readiness assessment.
\end{promptbox}

\begin{promptbox}{Exploration Falser}
Input:
  {problem}
  {knowledge directory, when available}
  {strategy card, including executed code-probe result when available}
  {previous proof and verifier feedback, when available}
  {falser reports inherited from child nodes, when available}

Attack the supplied strategy card. Do not rewrite, polish, or defend it,
except to name a minimal weakening. Examine the primary route first and
then each backup route.

Test the weakest central lemmas, hidden assumptions, overstrong claims,
boundary or counterexample regimes, dropped objections, theorem
hypotheses, inequality directions, and any mismatch between a code probe
and the claim it is said to test. For each serious issue, explain how the
route could fail and give a cheap decisive test when possible. A failed
program is missing evidence, not a mathematical counterexample; reserve
falsified for a valid argument or reproducible test.

Return an overall assessment, categorized objections, cheap falsification
tests, unresolved objections that later stages must preserve, minimal
weakenings to try, and a verdict indicating whether the route survives,
requires weakening or repair, should be rejected, or has been falsified.
\end{promptbox}

\begin{promptbox}{Exploration Aggregator}
Input:
  {problem}
  {knowledge directory, when available}
  {(strategy card, falser report) pairs sampled at this tree node}
  {previous strategy, falser report, and readiness-gate report, when
   available}
  {previous proof and verifier feedback, when available}

Produce one new strategy card, not a list and not a decomposition. Do not
average the children or merely select one of them. Preserve the narrowest
concrete, repairable route as the single primary route unless its core
mechanism has been killed. Retain at most two nonduplicative backup routes,
including a useful minority route when warranted.

Address every serious falser objection: reject, weaken, or repair the
attacked claim; refute the objection with a specific mathematical reason;
or carry the unresolved objection forward. Preserve unresolved blocking
obligations without recursively retelling the full history. Normalize
notation, preserve both branches of conditional gateways, and treat
computational results only at the strength actually tested.

Return a fresh strategy card together with an updated readiness judgment.
A route with an unresolved fatal obligation is not ready for decomposition.
\end{promptbox}

\begin{promptbox}{Readiness Gate}
Input:
  {problem}
  {knowledge directory, when available}
  {final strategy card}
  {attached falser report, when available}
  {previous proof and verifier feedback, when available}

Do not solve the problem or repair the strategy. Decide whether the card
is safe to turn into a proof decomposition. Audit every fatal bridge
claim, cited theorem and hypothesis, target equivalence, induction or
construction invariant, compatibility condition, hidden hard step,
bound direction, unresolved falser objection, and central code result.

Classify the strategy as ready for decomposition, eligible for decomposition
with explicit obligations, in need of further exploration, or to be rejected.

A strategy with an unresolved fatal bridge cannot be marked ready. It may
still proceed to decomposition, but only if the route and proof architecture
are stable and every remaining obligation has a concrete proof or
verification path. Treat these as minor obligations and assign each to an
explicit proof or certificate section in the proof plan.

An obligation is major if resolving it could require changing the route,
reduced target, or core mechanism, or if it has no clear repair path.
In such cases, return the strategy for further exploration.

Return the blocking issues, fatal bridge claims, theorem and criterion
audits, hidden hard steps, minimal remaining obligations, obligation
severity, whether decomposition is allowed, and the recommended next action.
\end{promptbox}

\subsection{Decomposition into Subproblems}
\label{app:prompt-decomposition}

The decomposer turns an approved strategy into a LaTeX skeleton and a DAG of
section-level subproblems. Each subproblem lists its earlier dependencies,
allowing eligible independent sections to be solved in parallel.

\begin{promptbox}{Decomposer}
Input:
  {problem}
  {knowledge directory, when available}
  {approved strategy card}
  {falser and readiness-gate reports}
  {previous decomposition, proof, and verifier feedback, when available}

Break the original problem into a short list of subproblems in topological
order. For each subproblem, list the indices of all earlier subproblems on
which it depends. Each subproblem should be mathematically meaningful,
nontrivial, and useful for reaching the final proof. Avoid tiny bookkeeping
steps.

Produce a standalone LaTeX master-document skeleton with a title, a
concise abstract describing the proof plan, and one section per
subproblem. Inside each section, write a short, mathematically specific
roadmap and reserve a unique location for the future section body.

Return the master-document skeleton together with the same ordered list
of subproblem titles, tasks, and dependency indices. Do not silently build
on unresolved attacked claims. Isolate them as explicit subproblems, weaken
them, or route around them.
\end{promptbox}

A decomposition aggregator combines candidate skeletons and their subproblem
DAGs into one plan. Unresolved bridge lemmas and certificate obligations must
remain explicit in that plan.

\subsection{Subproblem Solving and Aggregation}
\label{app:prompt-subproblems}

For each subproblem, multiple solvers produce candidate sections, paired falsers
critique them, and aggregators combine the candidates through a tree. On retry,
the previous attempt and its criticism are supplied so useful progress is
retained without hiding unresolved objections.

\begin{promptbox}{Subproblem Solver}
Input:
  {problem and assigned subproblem}
  {knowledge directory, when available}
  {current master document}
  {previous proof and global verifier feedback, when available}
  {previous attempt and attached criticism, on retry}
  {executed code-probe result, when available}

Write the full LaTeX body that replaces this section's placeholder. Do
not write a standalone document or repeat earlier sections. You may rely
on prior sections only to the extent justified by their text. Address
every relevant error already identified by global feedback or a previous
falser. If the section remains incomplete, mark the gap explicitly.

State exactly what theorem, lemma, reduction, or partial result the
section establishes. Separately identify the bridge claims on which it
depends and every obligation that remains unproved. Mark the section as
solved, partial, or blocked, and request a retry only for a concrete
remaining defect.

Only an executed code probe counts as computational evidence. If a
necessary certificate has not been executed, state only the reduction,
conditional result, or certificate plan and retain the stronger claim as
an open obligation.
\end{promptbox}

\begin{promptbox}{Subproblem Falser}
Input:
  {original problem}
  {assigned subproblem}
  {knowledge directory, when available}
  {current master document}
  {proposed section solution}
  {previous falser report, when available}

Do not repair or rewrite the section. Decide whether it actually proves
the assigned subproblem. Focus first on a fatal bridge: a false
equivalence, wrong target, wrong-sided bound, dropped factor, missing
constraint, index mismatch, unsupported theorem use, or code-probe
misuse. Audit every explicit open obligation and preserve earlier
objections unless the new section resolves them.

Classify the section as ready, conditional, or rejected. Ready means
later assembly may rely on the claimed result. Conditional means useful
progress remains but an explicit bridge or obligation is unresolved.
Reject a section that proves the wrong target or relies on a false or
constraint-dropping bridge. Any explicit gap precludes a ready verdict.

Return a concise summary, the fatal objections, and the claims that later
stages must not treat as established.
\end{promptbox}

\begin{promptbox}{Subproblem Aggregator}
Input:
  {problem and assigned subproblem}
  {knowledge directory, when available}
  {(section solution, falser report) pairs for one subproblem}
  {current master document}
  {previous proof and global verifier feedback, when available}

Synthesize the strongest single section replacement. Consume both each
candidate solution and its attached falser report. Do not silently
inherit an attacked claim: reject it, weaken it, repair it, or retain it
as an open obligation.

The resulting section must remain compatible with the notation,
assumptions, definitions, and established claims in the already-filled
master document. It must not contain a section heading or document
preamble. Preserve incomplete but useful progress only with an explicit
gap and a status of partial or blocked.

Return the synthesized section together with its claimed result, essential
bridge claims, unresolved obligations, completion status, and any concrete
reason that another attempt is needed.
\end{promptbox}

\subsection{Global Verification and Revision}
\label{app:prompt-global-verification}

The global verifier reviews the assembled proof together with compact
solver--falser audit cards. If the verifier rejects the proof, its feedback
guides targeted outline or section revision, after which the proof is verified
again.

\begin{promptbox}{Global Verifier}
Input:
  {original problem}
  {knowledge directory, when available}
  {complete assembled proof}
  {subproblem solver-falser audit cards}

Evaluate the proof against the original problem, not merely for internal
consistency. Read the entire proof from beginning to end even if a fatal
gap appears early. Check the exact hypotheses, scope, and conclusion, and
audit theorem uses, citations, computations, boundary cases, and every
claim attacked or left conditional in a subproblem audit card. A ready
local verdict is not itself a proof.

If the proof is complete, accept it and preserve the complete document.
Otherwise reject it and write a standalone, actionable feedback document
that localizes every material defect.

Verifier code probes are adversarial checks only. Use them to test a
claimed identity, bound, computation, or boundary case, never to repair
the proof.
\end{promptbox}

\begin{promptbox}{Revision-Specific Instructions}
Input:
  {original problem}
  {knowledge directory, when available}
  {current proof, decomposition, and ordered subproblems}
  {global verifier feedback}
  {original section and task, for section revision}

First identify every concrete mathematical issue in the global review.
For the outline cleanup, the allowed changes are to rewrite abstract or
roadmap material, slightly revise section names and tasks, and delete
redundant, empty, harmful, or superseded sections. Do not add sections,
invent placeholder tokens, reorder the surviving plan, or introduce a
new proof strategy.

For a section revision, fix every review issue that affects that section.
Keep material that is already correct and do not rewrite merely for
style. Preserve the paragraph order, displayed equations, named
environments, and LaTeX labels unless the review gives a concrete
mathematical reason to change them. Mark any remaining gap explicitly.

Return either a conservatively revised outline or a revised section body,
as appropriate. Candidate revisions are aggregated before the proof is
reassembled and verified again.
\end{promptbox}

\end{document}